\documentclass[sigconf]{acmart}
\usepackage{tikz}
\usetikzlibrary{shapes.geometric, arrows.meta, positioning, decorations.pathreplacing, calc}
\usepackage{algorithm}
\usepackage{algorithmic}
\usepackage{booktabs} 
\usepackage{siunitx}
\usepackage{stfloats}

\AtBeginDocument{%
  }

\begin{document}

\title{LithoFormer: A Robust Framework for Stratigraphic Inference via Transformers}

\author{Shwetha Salimath}
\affiliation{%
  \institution{Laboratoire Interdisciplinaire des Sciences du Numérique, \\ Université Paris-Saclay,\\ SLB }
  \city{Montpellier}
  \country{France}}
  \email{shwetha.salimath@universite-paris-saclay.fr}

\author{Francesca Bugiotti}
\affiliation{%
  \institution{Laboratoire Interdisciplinaire des Sciences du Numérique, CentraleSupélec, CNRS, Université Paris-Saclay}
  \city{Gif-sur-Yvette}
  \country{France}}
\email{francesca.bugiotti@centralesupelec.fr}

\author{Sylvain Wlodarczyk}
\affiliation{%
  \institution{SLB}
  \city{Montpellier}
  \country{France}}
\email{swlodarczyk@slb.com}

\author{Sohaib Ouzineb}
\affiliation{%
  \institution{SLB}
  \city{Montpellier}
  \country{France}}
\email{souzineb@slb.com}

\renewcommand{\shortauthors}{Salimath et al.}

\begin{abstract}

Accurate geological characterization of subsurface reservoirs from well log data is essential to support projects such as carbon capture and storage (CCS), geothermal development, and extraction of natural resources. Existing automated techniques for geological characterization primarily use sliding-window classification,  which limits their ability to understand broader geological contexts, often leading to misaligned formation layers.
To overcome these limitations, we introduce LithoFormer, a robust framework for stratigraphic inference using a Seq2Seq transformer model that ingests entire multivariate well logs in a single pass.  The framework utilizes a channel-independent PatchTST backbone enhanced with rotary positional embeddings (RoPE) to capture long-range geological dependencies across entire multivariate well logs. A decoupled multi-task head is employed to jointly predict geological zonation and precise boundary probabilities, while a geology-informed loss function enforces physical constraints such as the Law of Superposition. Validated and deployed on three real-world datasets, LithoFormer demonstrates a 90\% reduction in median boundary error and eliminates stratigraphic order violations compared to traditional sliding-window baselines. It also achieves a 80\% reduction in manual expert labor and eliminates stratigraphic inconsistencies, providing a scalable and reliable solution for large-scale subsurface modeling.
\end{abstract}



\maketitle

\section{Introduction}

The mapping of stratigraphy is essential for energy projects that require subsurface characterization, as it determines resource storage capacity and geological safety. Accurate 3D models of subsurface architecture are crucial for identifying porous reservoirs for carbon sequestration, geothermal formations, and traditional petroleum reserves. This modeling involves reconstructing geological processes over millions of years and primarily relies on well correlation, which aligns stratigraphic boundaries or "markers" across drilled wells \cite{tsuji2014reservoir, alam2023carbon}. Well logs analysis helps pinpoint these markers~\cite{darling2005well, ellis2007well}, providing a continuous, high-resolution record of subsurface layers often missed by seismic data.

Traditionally, well correlation using well logs is a manual process in which geologists visually match patterns in well log data, such as Gamma Ray, Resistivity, and Density measurements \citep{mann1978quantitative}. This method lacks scalability. It is confirmed by both geologists and our research that a specific marker within a region usually exhibits a consistent signal pattern \cite{rider1990gamma, abdel2022neural}. Early computational approaches for matching well log patterns, such as dynamic time warping (DTW) \cite{DTW}, provided some level of automation, but were often fragile, sensitive to noise, and struggled to generalize across geologically diverse areas. This led to the development of deep learning models that treated the task as a sliding-window classification problem. 
However, the sliding-window paradigm is inherently limited by a local receptive field. These models lack global context; a sliding window has a "short-sighted" view, unable to see the full geological sequence. They cannot distinguish between a valid marker pattern and a visually similar artifact appearing at an incorrect position, nor can they enforce the basic physical law that geological layers are deposited sequentially.

\begin{figure}[h!]
    \centering
    \includegraphics[width=\linewidth]{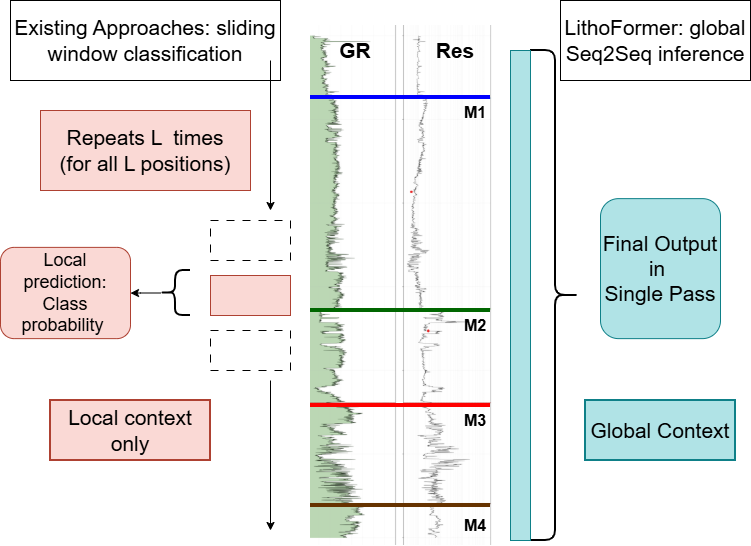}
    \caption{Workflow Comparison. The sliding-window approach (left) requires \(L\) local classifications per well, while LithoFormer (right), processes the entire well log in a single pass to generate a complete stratigraphic map. M1, M2, M3, and M4 are markers denoting the stratigraphic sequence.}
    \label{fig:conceptual_diagram}
\end{figure}

To address these challenges, we propose an approach shift from a local classification task to a global end-to-end Seq2Seq task. We introduce LithoFormer, whose core contribution is the ability to predict a sequential order of events from long, multivariate time series data. As illustrated in Figure \ref{fig:conceptual_diagram}, this approach uses the power of transformers to learn both local patterns and global trends that match a valid stratigraphic succession. By processing the well log in a single pass, LithoFormer ensures that every predicted marker is placed within its proper global context, maintaining the chronological integrity of the geological interpretation. The contributions of the paper are as follows.
\begin{itemize}
\item \textbf{LithoFormer Framework} for identifying a precise, ordered sequence of events within complex, multivariate time series data.
\item \textbf{A robust data-centric pipeline} that uses resampling, filtering, and constrained data augmentation to help the model distinguish true markers from out-of-context patterns.
\item \textbf{Multi-Task Transformer Architecture} using a channel-independent PatchTST transformer \citep{Yuqietal-2023-PatchTST}, enhanced with RoPE \citep{su2024roformer} and a decoupled multi-task head. The dual head jointly learns coarse zonations (geological formation intervals) and precise edges (interval boundaries).
\item A \textbf{Geology-Informed Loss} to enforce stratigraphic order (the Law of Superposition), while a \textbf{thickness-weighted} loss enables the model to detect extremely thin or rare layers accurately.
\item \textbf{Real World Validation}: We deploy and test on three datasets from Colorado, Wyoming, and the Norwegian North Sea. 
\end{itemize}

LithoFormer achieves state-of-the-art precision while ensuring geological consistency. This provides a scalable and physically consistent solution for subsurface reservoir characterization. 

The paper is structured as follows. In Section \ref{sec:related_work}, we review related work. Section \ref{sec:method} details our methodology, including the novel LithoFormer model architecture, the data augmentation pipeline, and the multi-stage training curriculum. Section \ref{sec:Results} reports our comprehensive experimental results and ablation analyses,  and concludes in Section \ref{sec:conclusion}.

\section{Related Work}
\label{sec:related_work}

Our work intersects three key research areas: automated geological correlation, deep learning for time series, and physics-informed machine learning. We review the evolution of well correlation and discuss advancements in time series transformers that support our approach.

\textbf{Automated Well Log Correlation} is essential in stratigraphic analysis. Initial methods focused on automation through signal processing, using cross-correlation, statistical methods, and feature-based matching \citep{dashtian2011analysis}. Techniques like DTW were used for aligning log sequences via optimal nonlinear warping \citep{lineman1987well}. However, DTW struggles with multi-well and multivariate correlation due to its computational demands and sensitivity to noise. Recent approaches have integrated machine learning (ML) techniques for better correlation. Recurrent Neural Networks (RNNs) and convolutional neural networks (CNNs), including hybrid models like LSTM-CNN and LSTM-2dCNN, have been successfully employed to process well logs as multivariate time series and identify marker signatures \citep{mlwellcorrelation, imamverdiyev2019lithological, salimath2025geots}. Nonetheless, challenges persist in managing long-range dependencies and capturing a well's global context, problems that transformer architectures aim to address.

\textbf{Transformers for Time Series Analysis.} 
The transformer architecture \citep{vaswani2017attention} has been adapted for time series tasks, particularly with models like Informer and Autoformer \citep{zhou2021informer, wu2021autoformer}, which focus on long-range forecasting. However, these models treat time series as sequences of individual time steps, leading to high computational costs. An advancement in this field is PatchTST \citep{Yuqietal-2023-PatchTST}, our architectural backbone, which segments a time series into patches used as tokens. This approach reduces sequence length, captures local semantics, and improves learning from longer contexts. Its has achieved state-of-the-art performance in classification \citep{wang2024medformer} and forecasting benchmarks \citep{huang2024long, goswami2024moment}. 
Transformers need positional information to model sequential dependencies effectively \citep{kimppt, dufter2022position}. Early methods employed absolute sinusoidal embeddings, which struggle with longer sequences. RoPE improves this by representing relative positions within the attention mechanism, enhancing translation invariance and generalization \citep{su2024roformer}. RoPE has shown exceptional performance in large language models like LLaMA and Gpt-oss \citep{touvron2023llama, gptoss}.

\textbf{Physics-Informed and Constrained Deep Learning.}
An important direction in scientific machine learning is to incorporate domain knowledge and physical principles directly into the learning process \citep{MLphysicsinformed, PINNai5030074}.
While initially introduced to solve physics problems, it was also successfully applied in finance and autonomous systems for fairness and safety \citep{Monotonic_neurips20, lossshapingconstraintslongterm}. Following this idea, our method introduces a loss term that enforces the geological law of superposition, guiding optimization toward stratigraphic sequences that are geologically consistent.

\section{LithoFormer framework}
\label{sec:method}

\begin{figure}[h]
    \centering
    \includegraphics[width= \linewidth]{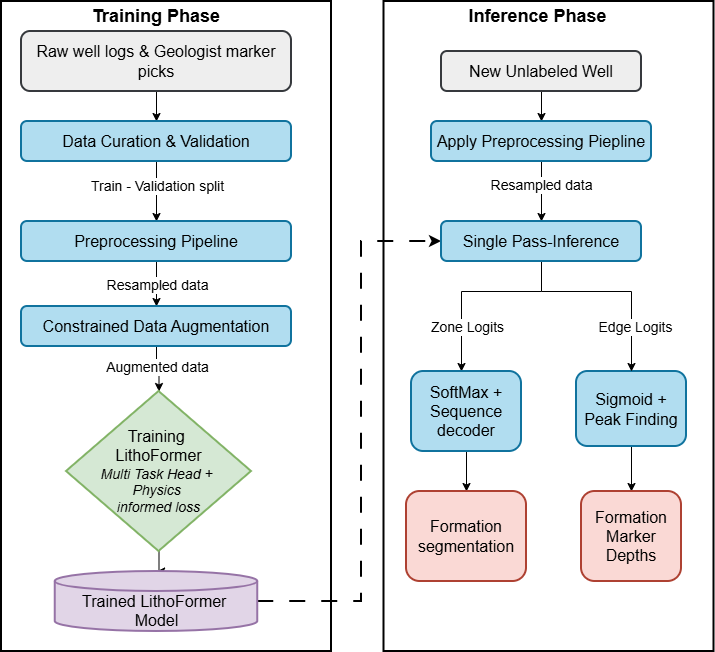}
    \caption{The end-to-end LithoFormer framework.}
    \label{fig:full_workflow}
\end{figure}

This section details the LithoFormer framework. We first explain the problem formulation followed by detail description of the data-centric pipeline, the multi-task architecture, and the geology-informed training curriculum. The complete end-to-end workflow, encompassing both training and inference, is illustrated in Figure \ref{fig:full_workflow}.

\subsection{Problem Formulation}
\label{sec:problem_formulation}

Stratigraphic inference is formulated as a mapping from a multivariate well log $X \in \mathbb{R}^{L \times C}$ to an ordered sequence of geological zones ${y}\in\{0,\dots, K\}^L$ with corresponding marker depths ${d}$ $\in \mathbb{R}^K$. Here, $L$ represents the number of depth samples and $C$ the number of well logging channels. These channels correspond to well log data of physical properties such as gamma-ray , resistivity, and density. $K$ is the number of markers. Zone indices $\{0, \dots, K\}$ correspond to markers sorted by expected stratigraphic order. The index "0" represents the zone before the first marker. Given  well $X$, LithoFormer is trained to simultaneously predict two outputs:
\begin{enumerate}
    \item \textbf{Zonation Probabilities $\hat{P}_{zone}$:} A set of probability curves  $\hat{P}_{zone} \in [0, 1]^{L \times (K+1)}$ indicating the probability of geological zones occurrence at each depth.
    \item \textbf{Edge Probabilities $\hat{P}_{edge}$:} A set of probability curves, $\hat{P}_{edge} \in [0, 1]^{L \times K}$, indicating the probability of marker occurrence at each depth.
\end{enumerate}

A globally consistent zone sequence $\hat{y}_{zone} \in \{0,\dots,  K\}^L$ is derived from $\hat{P}_{zone}$ to establish the stratigraphic order. In parallel, precise marker depths $\hat{d}_{edge} \in \mathbb{R}^K$ are computed by identifying peak probabilities within $\hat{P}_{edge}$. With this notation, the fundamental geological constraints described in the paper can be stated as:
\begin{itemize}
\item \textbf{Monotonic layer order}: For any two depth indices $i < j$, the predicted zone indices must be increasing: $\hat{y}_i \le \hat{y}_j$.
\item \textbf{Exactly one boundary per marker}: For each marker $k \in \{1,\dots,  K\}$, $\hat{P}_{edge}$ should contain at the $k$-th channel a single dominant peak at $\hat{d}_{edge,k}$. This reflects the geological assumption that marker $k$ occurs exactly once and in the correct order.
\end{itemize}

\subsection{Data-Centric Pipeline}
\label{sec:data_pipeline}

Real-world well log data often contain sensor noise, data gaps, and inconsistent sequence lengths due to varying drilling depths. To manage this variability, a preprocessing pipeline is used. First, the logs are resampled to a fixed sequence length. Then, a two-stage filtering process is applied: a Hampel filter \citep{pearson2016generalized} replaces outliers with the local median, and a Savitzky-Golay filter \citep{schafer2011savitzky} smooths the signal while preserving the trend. To combat overfitting and ensure accurate depth predictions, we propose a constrained data augmentation strategy that inserts expert-verified marker signatures into incorrect log locations, generating realistic out-of-context negative samples. Our ablation studies \ref{sec: ablation} demonstrate that this method significantly improves model robustness and precision.

From a mathematical point of view, let $X \in \mathbb{R}^{L \times C}$ be the input well log and $y \in \{0, \dots, K\}^L$ its target zonation sequence. Let $\mathcal{M}_y$ be the set of true markers present in $y$. We generate a new sample $(X_{aug}, y)$ as follows:
First, a true marker $m^* \in \mathcal{M}_y$ with its corresponding depth index $d^*$ is selected. Extract the multivariate signature pattern $X_{sig} \in \mathbb{R}^{2W \times C}$ from a window of length $2W$ samples around $d^*$.
To prevent overlaps with existing markers, we define $\mathcal{Z}_{exclude}$ as the union of windows of size $2E$ samples around all marker depth indices in $\mathcal{M}_y$.
Sample a new depth index $d_{new}$ from the set of valid indices $\{1, \dots, L\} \setminus \mathcal{Z}_{exclude}$.
Finally, construct $X_{aug}$ by copying $X$ and pasting the pattern $X_{sig}$ at the new location $d_{new}$. The new pair $(X_{aug}, y)$ is then added to the training set.

\subsection{LithoFormer Transformer Backbone}
\label{sec:architectural_components}

LithoFormer employs a multi-task Seq2Seq transformer backbone to process multivariate well log inputs \( X \in \mathbb{R}^{L \times C} \) through two sequential blocks: a channel-independent feature extraction PatchTST encoder and a multi-task prediction head. The backbone treats each log channel as a univariate series, which is essential for two reasons. First, it accommodates diverse statistical distributions and log patterns of different physical measurements. Second, it ensures robustness against missing data. While Gamma Ray (GR) is often available, other logs such as resistivity (RES) or density (DEN) are frequently absent. Channel independence allows effective feature extraction from available signals without interference from missing inputs.

\begin{figure*}[h!]
    \centering
    \includegraphics[width=0.85\textwidth, height = 5 cm]{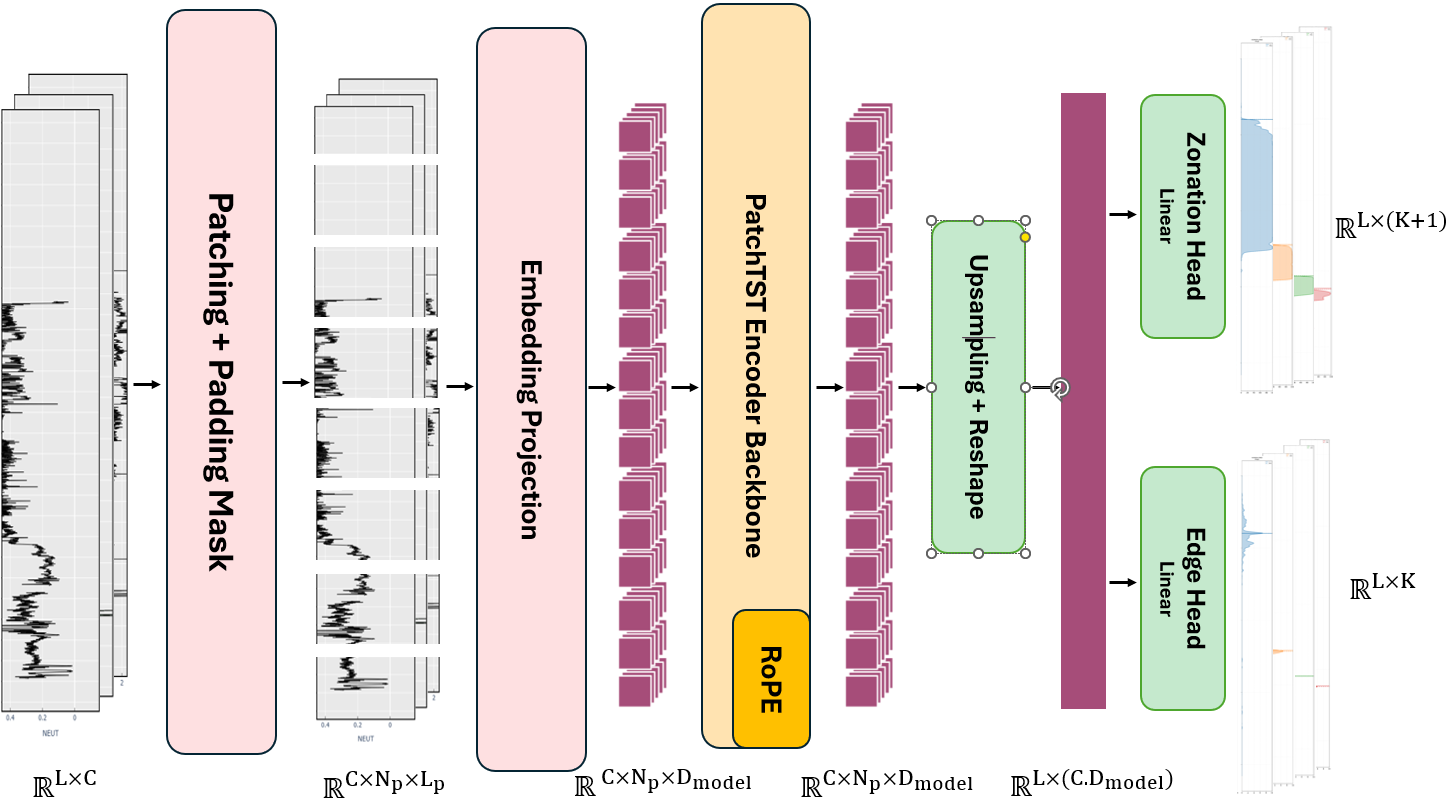} 
    \caption{The LithoFormer transformer backbone. Multivariate inputs are processed via a channel-independent PatchTST with RoPE. Decoupled multi-task heads then generate zone and edge probabilities from the fused dense features.}
    \label{fig:architecture}
\end{figure*}

The input data is segmented into \(N_p\) overlapping patches of length \(L_p\) along the depth dimension with patch stride \(s\), processed per input channel due to the channel-independent architecture. These patches are projected onto a latent dimension \(D_{\text{model}}\), and the resulting latent embedding vectors are processed by a PatchTST encoder. To effectively capture the relative depth dependencies crucial to stratigraphy, where sequence order and layer thickness are important, RoPE is applied. RoPE embeds relative positional information directly into the attention mechanism, enhancing translation invariance and generalization to longer sequences \citep{su2024roformer}.

 To upsample compressed patch features of the encoder output back to a dense well log resolution \(L\), linear interpolation is applied. Following the upsampling procedure, we obtain a tensor in \(\mathbb{R}^{C \times L \times D_{\text{model}}}\). Subsequently, the feature vectors corresponding to all \(C\) logs are concatenated to construct a dense unified representation \(Z_{\text{upsampled}} \in \mathbb{R}^{L \times (C \cdot D_{\text{model}})}\). This operation facilitates the learning of cross-log dependencies prior to the prediction head.

As an innovative aspect of our model, we utilize two decoupled prediction heads running in parallel to forecast zonation and edge probabilities. The zonation head focuses on the segmentation task by projecting the latent features into \(K+1\) channels via a linear layer, which corresponds to the geological zone probabilities \(\hat{P}_{zone}\) at each depth index. This approach effectively captures the global structural state of the stratigraphy. In parallel, the edge head processes the latent features to predict marker depth probabilities \(\hat{P}_{edge}\) through a linear layer. Unlike the zonation head, which predicts mutually exclusive zones, the edge head is designed to identify sparse boundary events by generating independent logits for each of the \(K\) specific markers.
The output $\hat{P}_{zone}$ is as follows:
\begin{equation}
\hat{P}_{zone} = \text{Linear}_{zone}(Z_{upsampled}) \in \mathbb{R}^{L \times (K+1)}
\end{equation}
The output $\hat{P}_{edge}$ is as follows:
\begin{equation}
\hat{P}_{edge} = \text{Linear}_{edge}(Z_{upsampled}) \in \mathbb{R}^{L \times K}
\end{equation}

A key feature of LithoFormer is the joint optimization of both heads without gradient isolation. This allows the model to reconstruct an ordered sequence of events while preserving spatial detail through shared gradient backpropagation.

\subsection{Geology-Informed Loss Function}
\label{sec:loss_function}

Standard cross-entropy loss is insufficient for our task because it treats each depth index independently and ignores stratigraphic sequence order. To obtain geologically plausible predictions, we design a composite loss that embeds domain knowledge by combining the cross-entropy term ($\mathcal{L}_{CE}$) with regularizers. These regularizers preserve stratigraphic order ($\mathcal{L}_{Mono}$) and sharpen edge detection boundaries ($\mathcal{L}_{Edge}$) as observed in ablation studies \ref{sec: ablation}. The total loss, \(\mathcal{L}_{\text{total}}\), is a weighted sum of these components, with $\lambda_m, \lambda_e \in \mathbb{R}$.

\begin{equation}
\mathcal{L}_{total} = \mathcal{L}_{CE} + \lambda_m \mathcal{L}_{Mono} + \lambda_e \mathcal{L}_{Edge}
\end{equation}

\textbf{Weighted Cross-Entropy ($\mathcal{L}_{CE}$):} Geological layers vary in thickness, leading to class imbalance in the zonation sequence, and as a result, thin yet significant formations may be overlooked. Therefore, we use the inverse thickness weighting, where the weight $w_k$ for each zone class $k$ is $ L \div N_{k}$, where \( N_k \) is the number of samples inside the corresponding zone.

\textbf{Monotonic Loss ($\mathcal{L}_{Mono}$):} An important physical constraint is the enforcement of the Law of Superposition, which requires that stratigraphic sequences be arranged by geological age from younger to older strata with increasing depth. The expected zonation class sequence $\hat{k}_{zone}$ is calculated as $\hat{k}_{zone, i} = \sum_{k=0}^{K} k \cdot \hat{P}_{zone,i,k} $ at each depth index $i$. A penalty is applied only when the $\hat{k}_{zone}$ sequence violates increasing monotonicity: 

\begin{equation}
\mathcal{L}_{Mono} (\hat{k}_{zone}) = \frac{1}{L-1} \sum_{i=2}^{L} \text{ReLU}(\hat{k}_{zone, i-1} - \hat{k}_{zone, i})
\end{equation}

\textbf{Edge Focal Loss ($\mathcal{L}_{Edge}$):} As the number of zonation boundaries is much fewer than the total number of depth samples, it creates a significant class imbalance. Let ${y}_{edge} \in \mathbb{R}^{L\times K}$ be encoded by the marker channel as a Gaussian distribution centered around the true marker depth index.
To obtain $\mathcal{L}_{Edge}$, we first define the focal loss $\mathcal{L}_{\text{FL,i,k}}$ for a depth index $i$ and a marker class $k$. The focal loss is derived from the the Binary Cross-Entropy loss $\mathcal{L}_\text{BCE,i,k}$ applied to the edge head output probabilities $\hat{P}_{edge}$ against the target ${y}_{edge}$ at the depth index $i$ and marker channel $k$ such that:
\begin{equation}
\mathcal{L}_{\text{FL,i,k}} = \alpha_{i,k} (1 - \exp{(- \mathcal{L}_{\text{BCE,i,k}}}))^\gamma \cdot \mathcal{L}_{\text{BCE,i,k}}
\end{equation}
 The focusing parameter \( \gamma \) decreases $\mathcal{L}_{\text{FL,i,k}}$ for correctly classified examples. Let $\alpha \in [0.5,1]$, the balancing factor \( \alpha_{i,k} \) is defined as $\alpha$ when ${y}_{edge, i, k} > 0.5$ and $1 - \alpha$ otherwise. 
 Higher weights are assigned to indices near true marker depths, and lower weights to background indices. 
$\mathcal{L}_{Edge}$ is the mean focal loss over all depth indices and marker classes. This loss term is crucial for achieving high precision in sparse detection tasks.

\subsection{Stratigraphic Inference}
\label{sec:inference}

The LithoFormer framework generates two complementary sets of predictions: $\hat{P}_{zone}$ and $\hat{P}_{edge}$. We get $\hat{d}_{zone}$ by employing a sequence decoder that iteratively selects for each $\hat{P}_{zone}$ channel the minimal depth at which the corresponding probability is the highest across channels. Concurrently, high-resolution marker depths ($\hat{d}_{edge}$) are obtained by identifying the peak probabilities within each marker channel of the output $\hat{P}_{edge}$. $\hat{y}_{zone}$ and $\hat{y}$ is the zone sequence representation of $\hat{d}_{zone}$ and $\hat{d}_{edge}$. Ultimately, the framework utilizes the $\hat{d}_{edge}$ as the final result, leveraging their superior local precision while the zonation output $\hat{y}_{zone}$ serves to validate the global stratigraphic sequence.

\section{Experiments and Results}
\label{sec:Results}

We validated the LithoFormer framework and our backbone model by comparing its performance in precision, recall, and stratigraphic order against state-of-the-art methods using an end-to-end Seq2Seq approach and sliding-window models. Ablation studies explored our data-centric pipeline, patch stride, and loss function terms. The following sections outline our experimental setup, results, and performance analysis.

\subsection{Datasets}
\label{sec:datasets}

To perform a robust evaluation, we use three distinct datasets from publicly available well-log data, representing various levels of geological complexity and signal degradation. We filter training wells, with at least 60\%  sequence markers present. A quality control step removes wells with entirely missing log readings and label quality control to remove marker outliers. We perform no imputation, smoothing, or input signal cleaning; thus, the model is evaluated on raw, noisy industrial data. The train, validation, and test split is 70:20:10. The  test set wells were manually verified by a domain expert to address inaccuracies in public datasets.

\begin{table}[h]
\centering
\setlength{\tabcolsep}{2pt}
\caption{Dataset Specifications}
\label{tab:datasets}
\begin{tabular}{l c c c c c}
\toprule
\textbf{Dataset} & \textbf{Wells} & \textbf{Interval (ft)} & \textbf{Channels} & \textbf{Missing} \\
\midrule
Colorado & 800 & 5,800--7,200 & 1 (GR) & 0\% \\
Wyoming & 700 & 100--1,800 & 1 (GR) & $\sim$3\% \\
North Sea & 1000 & 2,000--14,000 & 3 (GR, RES, DEN) & $>$54\%\textsuperscript{*} \\
\bottomrule
\multicolumn{5}{l}{\footnotesize \textsuperscript{*}Missingness in North Sea is primarily driven by Density (DEN) logs.}
\end{tabular}
\end{table}

The three datasets of increasing difficulty levels are used for model evaluation (Table \ref{tab:datasets}). Dataset Colorado \cite{colorado_data} provides a baseline where marker signatures for Niobrara, Codel, and Forth Hayes are clearly distinguishable. Dataset Wyoming \cite{wyoming_data} adds complexity with similar signature patterns in Badger Coal, Felix Coal, and Big George Coal markers, some lacking recognizable signatures. Dataset Norwegian North Sea \cite{npd_data} represents a complex, multivariate, long-sequence scenario critical for proving scalability. The zonation sequence applied is Hordaland, Rogaland, Shetland, Viking, Vestland, and Dunling.

\subsection{Experimentation setup}
\label{sec:baseline}

To evaluate LithoFormer, we benchmarked it against four baselines from different architectural paradigms. First, a traditional DTW-based baseline represents the industry standard. Second, GeoTS, our previous state-of-the-art model using the LSTM-2dCNN neural network \cite{salimath2025geots}. Third, a Conformer model was assessed for the combined CNN and transformer features. Fourth, a PatchTST baseline was included to evaluate performance gains from the LithoFormer backbone's multi-task head. Lastly, to validate our channel-independent backbone choice, we implemented a channel-mixing variant combining log channels at the embedding layer. To ensure a fair comparison, all neural baselines—including Conformer, PatchTST, and Channel-Mixing were integrated with LithoFormer framework's data-centric pipeline. All baseline models underwent rigorous hyperparameter tuning for a fair evaluation.

The models were implemented in PyTorch and trained on a NVIDIA L4 GPU using the fastai framework \cite{FastaiHoward_2020}. The transformer backbone features six PatchTST encoder layers with four attention heads, a patch stride $s$ of eight, and a patch length $L_p$ of 50. Data augmentation window parameters $W$ and $E$ are set to 100 ft and 200 ft, and adjusted to the resampling rate of the input well log sequence. The loss function is calibrated with \(\lambda_m = 1.5\) ($L_m$) and \(\lambda_e = 20.0\) ($L_e$). The code for reproducing the models and experiments will be made publicly available upon publication.

We evaluated all models using three critical metrics. Our primary metric for localization precision is the Median Absolute Error (MedAE), reported in feet for its robustness against outlier predictions in real-world datasets. To assess sensitivity, we used Recall@$\tau$, defined as the proportion of markers predicted within $\tau$ feet of the ground truth. Lastly, for geological consistency, we report the accuracy of the predicted sequence order, reflecting the percentage of wells where all predicted markers conform to the correct stratigraphic sequence.

\subsection{Main Results}
\label{sec: main_results}

The results in varying geological complexities are summarized in Table~\ref{tab:main_results}. A clear performance hierarchy emerges as data difficulty increases. In the Colorado dataset, transformer-based models excelled, achieving MedAE below 3 ft with perfect recall, indicating efficacy with simpler stratigraphic sequences. In the Wyoming dataset, the sliding-window approach showed limitations; the GeoTS baseline’s recall fell to 50.37\%, while patch transformer models retained accuracy, highlighting the advantages of a global context approach.

\begin{table*}[h]
\centering
\caption{Main comparative results across all datasets. Recall and order accuracy values are percentages (\%). MedAE is in feet.}
\label{tab:main_results}
\sisetup{table-align-text-post=false} 
\begin{tabular}{@{} l
                S[table-format=2.1] S[table-format=2.0] S[table-format=3.0] S[table-format=1.2]
                S[table-format=2.1] S[table-format=2.0] S[table-format=3.2] S[table-format=1.2]
                S[table-format=2.1] S[table-format=2.1] S[table-format=3.0] S[table-format=1.2] @{}}
\toprule
 & \multicolumn{3}{c}{\textbf{Dataset Colorado}} & \multicolumn{3}{c}{\textbf{Dataset Wyoming}} & \multicolumn{3}{c}{\textbf{Dataset North Sea}} \\
 \cmidrule(lr){2-4} \cmidrule(lr){5-7} \cmidrule(lr){8-10}
\textbf{Model} & {MedAE(ft)} & {Recall@20ft} & {Order \%} & {MedAE(ft)} & {Recall@20ft} & {Order \%} & {MedAE(ft)} & {Recall@50ft} & {Order \%} \\ \midrule
DTW Baseline & {6.60} & {89.25} & {100} & {194.0} & {15.14} & {57.8} & {1123} & {10.25} & {20} \\
GeoTS (KDD '25) & {2.0} & {97.42} & {80}  & {76.2} & {50.37} & \textbf{100} & {667.98} & {35.24} & {88} \\
Conformer$^*$ & {15.0} & {53.75} & \textbf{100} & {322.4} & {21.67} & {85} & {410.92} & {38.92} & {68} \\
Patchtst baseline$^*$ & {3.0} & {97.67} & \textbf{100} & {8.0} & {89.62} & {98}  & {84.67} & {51.56} & {92} \\
Channel-Mixing$^*$ & \textbf{2.0} & \textbf{100.00} & \textbf{100} & \textbf{2.0} & \textbf{91.62} & \textbf{100} & {59.92} & {58.54} & {94} \\ 
\textbf{LithoFormer (Ours)} & \textbf{2.0} & \textbf{100.00} & \textbf{100} & \textbf{2.0} & \textbf{91.62} & \textbf{100} & \textbf{19.74} & \textbf{72.81} & \textbf{95} \\ \bottomrule
\multicolumn{10}{l}{\footnotesize $^*$These baselines utilize the same data-centric pipeline as LithoFormer.}
\end{tabular}
\end{table*}

\begin{figure*}[h]
    \centering
    \includegraphics[width=0.98\textwidth, height = 6.5cm]{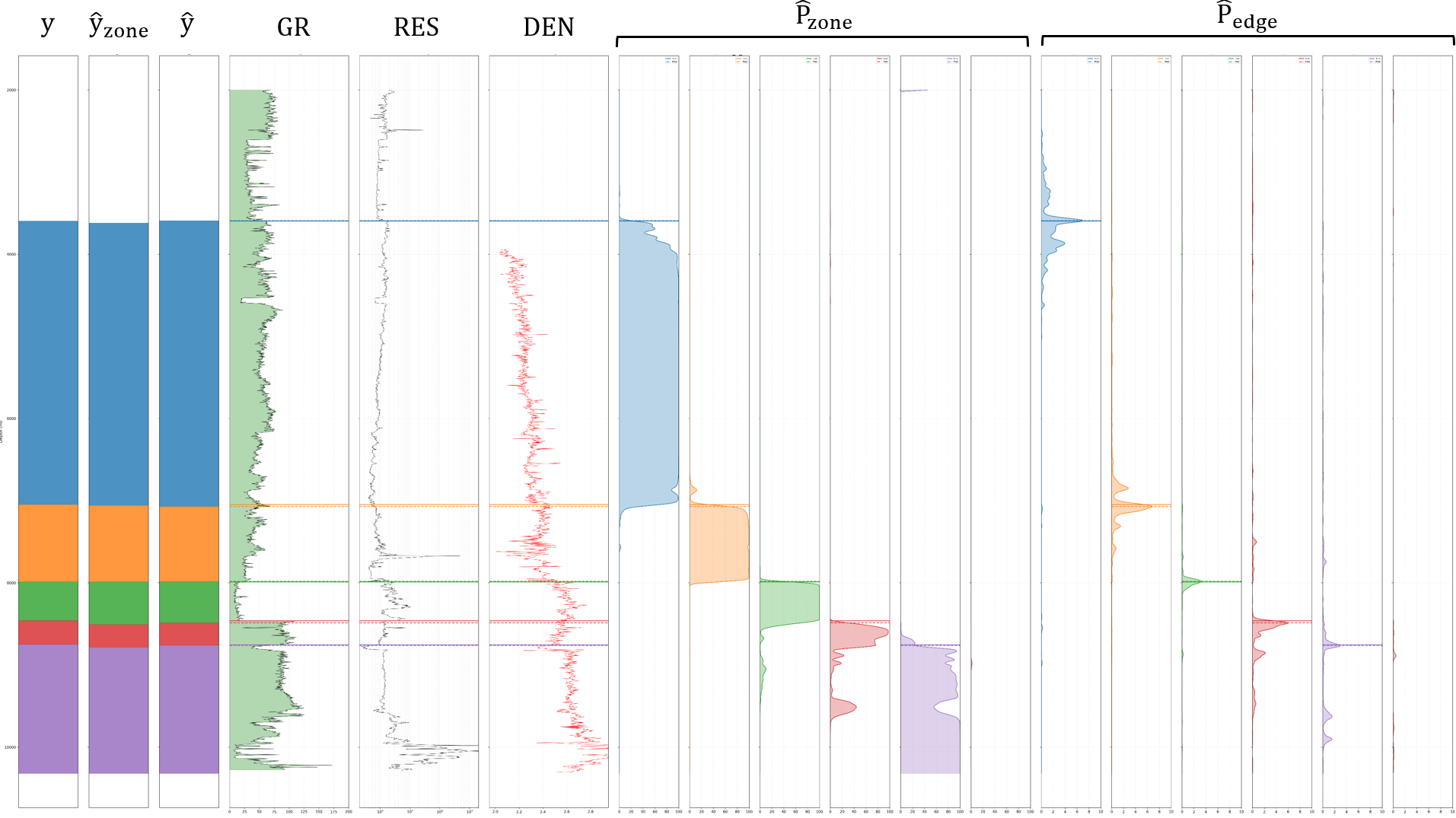} 
    \caption{Qualitative visualization of LithoFormer's output on a North Sea well. 
    \textbf{Tracks (1-3):} Ground Truth (Geologist picks), predicted Zone and Edge sequence. 
    \textbf{Tracks (4-6):} The multivariate input logs. 
    \textbf{ Tracks (7+):} The internal model outputs. The \textit{Zone Probabilities} identify the geological zone, while the \textit{Edge Probabilities} precisely pinpoint the boundary location.}
    \label{fig:qualitative_results}
\end{figure*}

The analysis of the North Sea dataset reveals critical insights. We report recall at \( \tau= 50 \) ft, considering a downsampling rate of 4 feet. The DTW baseline and GeoTS show limitations due to local matching issues, while the Conformer model performs suboptimally. In contrast, the channel-mixing transformer achieves a MedAE of 59.92 ft, but our channel-independent LithoFormer excels at 19.74 ft. This indicates that early fusion harms well log data analysis, as channels represent distinct properties with different noise profiles. LithoFormer demonstrates a significant advancement in the learning of temporal structures, achieving a new benchmark in error reduction. It decreases errors by 95\% compared to GeoTS and by 75\% against the PatchTST baseline. LithoFormer also achieves the highest order accuracy. Figure \ref{fig:qualitative_results} illustrates the inference with distinct geological zonation and precise boundary detection.

\noindent \textbf{Failure Analysis for North Sea} While \text{LithoFormer} shows high recall in Colorado and Wyoming, the North Sea (NS) dataset has a $27\%$ recall gap due to geological complexity. Key drivers for these stratigraphic residuals include \textit{Geological Ambiguity}, which arises from difficult-to-distinguish markers like the Hordaland Group. Additionally, \textit{Class Imbalance} plays a significant role, as there are under-represented formations such as the Vestland and Dunlin groups that have 75\% fewer samples. The presence of \textit{False Positives} is another contributing factor, stemming from bimodal probability peaks resulting from near-identical signatures. Furthermore, \textit{Sensor Noise} can lead to data gaps being misinterpreted as stratigraphic edges, and \textit{Covariate Shift} results in precision loss in deeper test wells. To address these challenges, the model flags low-confidence edges for manual expert review during production workflows.

\subsection{Ablation study}
\label{sec: ablation}

To analyze the key components of our framework and validate our design choices, we performed a series of component-wise ablation studies. They are conducted on the multivariate North Sea dataset, where the impact of each component is most pronounced. The results shown are on the 100 test wells.

\paragraph{Backbone Stride and Resource Efficiency}
\begin{table}[h]
\centering
\caption{Ablation on Backbone Stride (North Sea)}
\label{tab:ablation_stride}
\begin{tabular}{@{} ccccc@ {}}
\toprule
\textbf{Stride} & \textbf{MedAE} & \textbf{Recall} & \textbf{Run memory (GB)} \\ \midrule
\text{Default (1)} & 26.81 & 69.67 & 17.8 \\
4 & 22.74 & 69.20 & 1.8 \\
\textbf{8} & \textbf{19.74} & \textbf{72.81} & \textbf{0.84} \\
16 & 28.60 & 65.20 & 0.58 \\
\bottomrule
\end{tabular}
\end{table}

The trade-off between feature resolution and resource consumption was analyzed by adjusting the stride of the PatchTST backbone (Table \ref{tab:ablation_stride}). A stride of 1 captures fine temporal details but results in a high MedAE and memory usage. In contrast, Stride 8 achieved the best MedAE and Recall while also reducing GPU usage to 0.84 GB due to fewer tokens. This performance gain stems from coarser sampling, which reduces overfitting to high signal variability. Moreover, upsampling latent features via linear interpolation offers a more stable representation for prediction heads compared to the denser stride 1. The 95\% decrease in GPU memory usage to 0.84 GB enables deployment on edge hardware with less than 1GB of VRAM.

\paragraph{Loss Components}

\begin{table}[h]
\centering
\caption{Ablation on Loss Function Components (North Sea) }
\label{tab:loss_ablation}
\begin{tabular}{l c c c}
\toprule
\textbf{Loss Configuration} & \textbf{MedAE} & \textbf{Recall} & $\textbf{Order}$ \\
\midrule
1. Baseline ($\mathcal{L}_{CE}$) & 72.38 & 45.74 & 52.9 \\
2. $\mathcal{L}_{CE}$+ Physics ($\mathcal{L}_{Mono}$) & 58.80 & 49.39 & {80.57} \\
3. $\mathcal{L}_{CE}$ + Edge ($\mathcal{L}_{edge}$) & 25.85 & 65.88 & 73.0 \\
\textbf{4. LithoFormer ($\mathcal{L}_{total}$)} & \textbf{19.74} & \textbf{72.81} & \textbf{94.56} \\
\bottomrule
\end{tabular}
\end{table}
The contribution of each loss component was evaluated by isolating its effect (Table \ref{tab:loss_ablation}). The first and second rows, lacking an edge loss term, derive metrics using $\hat{d}_{zone}$, while the third and fourth use $\hat{d}_{edge}$. The baseline cross-entropy loss ($\mathcal{L}_{CE}$) shows poor precision and order due to a lack of geological context. Adding ($\mathcal{L}_{Mono}$) boosts the raw order to 80.57\%, reinforcing the Law of Superposition. Row 3 ($\mathcal{L}_{Edge}$) is vital for precision, significantly enhancing MedAE and Recall. In conclusion, LithoFormer ($\mathcal{L}_{total}$) achieves optimal performance by integrating structural zonation logic with edge detection.

\paragraph{Data-Centric pipeline}
Assessments of the data pipeline (Table \ref{tab:datapipeline_ablation}) showed that constrained data augmentation significantly improves model performance, achieving a 67\% reduction in MedAE and a 26\% increase in recall. Showing that learning the global stratigraphic context via cut-and-paste negative examples is important. However, explicit preprocessing with Hampel/Savitzky-Golay filters negatively impacted the final framework. While it slightly improved the baseline, it increased the MedAE of the augmented framework. The best approach is to use raw, resampled logs, allowing the multi-task architecture to perform internal denoising.
\begin{table}[h]
\centering
\caption{Ablation on Data-Centric Components (North Sea)}
\label{tab:datapipeline_ablation}
\begin{tabular}{@{} lccc@ {}}
\toprule
\textbf{Configuration} & \textbf{MedAE} & \textbf{Recall} & $\textbf{Order}$ \\ 
\midrule
Baseline (Raw Data) & 60.74 & 46.29 & 89  \\
Preproc & 55.03 & 47.89 & 89 \\
Constr Aug. (Final) & \textbf{19.74} & \textbf{72.81} & \textbf{94.56} \\
Preproc. + Aug. & 25.34 & 67.99 & \textbf{94.56} \\
\bottomrule
\end{tabular}
\end{table}

\section{Conclusions and Future Work}
\label{sec:conclusion}

This work introduces LithoFormer, a robust framework designed for high-precision stratigraphic inference from multivariate well-log data. By reframing the problem from traditional sliding-window classification to a global seq2seq paradigm, the framework successfully captures the long-range dependencies essential for reconstructing geologically consistent sequences. Key innovations include a decoupled multi-task architecture and a geology-informed training. Validated across three diverse datasets the LithoFormer demonstrates exceptional stability, maintaining high precision even in the presence of 54\% data missingness. Ultimately, LithoFormer transforms a traditionally slow analytical process into a scalable automatic solution, applicable beyond geology to fields like medical monitoring and industrial phase detection. 

Future work aims to develop a global foundation model for well log data using a self-supervised approach similar to the MOMENT model~\cite{goswami2024moment}. This model will support task-specific heads for data imputation, marker prediction, and petrophysical property estimation, enabling faster adaptation to new regions with fewer expert engagements.

\begin{acks}
We thank the CESMC \cite{colorado_data}, WOGCC \cite{wyoming_data}, and NOD \cite{npd_data} for providing well log datasets for informational and research purposes. 
\end{acks}

\bibliographystyle{ACM-Reference-Format}
\bibliography{simple}

\end{document}